\documentclass[10pt,twocolumn,letterpaper]{article}
\usepackage{wacv}
\usepackage{times}
\usepackage{epsfig}
\usepackage{graphicx}
\usepackage{amsmath}
\usepackage{amssymb}
\usepackage{tabularx}
\usepackage{multirow}
\usepackage{xcolor}
\usepackage{listings}
\def\wacvPaperID{477} 

\wacvfinalcopy

\ifwacvfinal
\def\assignedStartPage{1} 
\fi
\ifwacvfinal
\usepackage[breaklinks=true,bookmarks=false]{hyperref}
\else
\usepackage[pagebackref=true,breaklinks=true,colorlinks,bookmarks=false]{hyperref}
\fi

\ifwacvfinal
\else
\fi

\pdfoutput=1
\usepackage{algorithm, setspace}
\usepackage{algorithmic, setspace}

\begin{document}
\title{\bf Incorporating Each Layer Depth Information for Better Initialization Scheme }
\author{Vijay Pandey\\
IBM\\
India\\
{\tt\small vijay.pandey1@ibm.com}
}
\maketitle
\begin{abstract}
In past few years, various initialization schemes have been proposed. These schemes are glorot initialization, He initialization, initialization using orthogonal matrix, random walk method for initialization. Some of these methods stress on keeping unit variance of activation and gradient propagation through the network layer. Few of these methods are independent of the depth information while some methods has considered the total network depth for better initialization. In this paper, comprehensive study has been done where depth information of each layer as well as total network is incorporated for better initialization scheme. It has also been studied that for deeper networks theoretical assumption of unit variance throughout the network does not perform well. It requires the need to increase the variance of the network from first layer activation to last layer activation. We proposed a novel way to increase the variance of the network in flexible manner, which incorporates the information of each layer depth. Experiments shows that proposed method performs better than the existing initialization scheme.
\end{abstract}
\section{Introduction}
Initialization schemes has its own importance in neural networks. Initialisation scheme of the weights have enough scope in helping the network in proper training. Which initialisation scheme is opted affects the weight distribution in effective way. After introduction of batch, normalization in deep learning importance of initialization schemes has shown its little importance
It is worth noted fact that loss function of neural networks are not well understood until the time. Many researchers have explained the working of loss function for better generalization however better understanding is still need do explore. From where you start results in better convergence and better results on unseen data. Better initialization points is still not well understood nevertheless some promising work have been done in this area, which works well for some class of problems. It has been observed that two hidden units having same parameter and input value results in bad initialization scheme. In previous researches, initialization scheme does depend on number of input connections and output connections. Depth of the layer is not considered as part of the hyper-parameter. Natural question arises here, does depth not contribute significantly in parameters space. This question is answered in this paper. It has been analyzed in the paper that, early layers should carry strong signals compared to last layers. It is evident that last layers are trained based on the early layers. If early layers do carry weak signals to propagate further in the network then how does later layers can extract robust pattern out of it. Based on the analogy it is experimented that if depth is also considered while initializing then it helps in propagating the signals in appropriate way. Let us make some discussion on how things work in parameter space. While training, it is experienced that weights on the last layer does change in that quantity as early layers do not. This is due to the vanishing gradient problem where proper gradient information is not carried back to the earlier layers. If total network depth is $L$ and depth of the $i^{th}$ layer is $d_i$ . We scale the variance of each layers weight with factor $\alpha \geq 1$. To give more than unit variance to the network we studied three distinct approaches. In our discussion we will consider three cases; 1) Weights of all layer are initialized with constant factor. 2) Weights of each layer are initialized with increasing order of their depth. 3) Weights of each layer are initialized with decreasing order of their depth. In following subsection we discuss these approaches in more detail.
\section{Related work}
Numerous ground-breaking results have been produced in this direction. First significant result was given by glorot et. Al \cite{Glorot2010UnderstandingTD}, where they first shown the importance of initialization scheme on the training of the neural network. Experiment was done primarily for the sigmoidal activation function and it was shown that tanh performs better as compared to sigmoid function. It suggested that number of input neurons and output neurons of a given layer l participates in amplifying the activation and back propagation signal. Proof is given on variance-based method where target is to maintain unit variance of forward and backward signal across the network. In other commendable findings by he et.al \cite{He_init_2015}. It was noticed that previously proposed initialization scheme considers the sigmoidal activation funcion. However, in modern practice rectified linear units are largely used. Therefore, extending the original proof for Relu activation and facilitated with new initialization scheme. In   another research by orthogonal initialization \cite{saxe2013exact}. it has been concluded that  for some class of problems, novel orthogonal matrix will work well to initialize the weights. There is another popular paper where concept of random walk is used for better variance \cite{sussillo2014random}. Concept of orthogonal matrix and batch normalization is used to enhance the performance \cite{good_init}. Some other significant work is carried in \cite{Aguirre2019ANS}.
\section{Preliminaries}
\subsection{Notations}
$prodn(a, start, end) = \prod_{l=start}^{end} a_l$
\subsection{Forward Propagation Case}
In this sub-section, we will study the forward propagation step. We will proceed with the proof given in the \cite{Glorot2010UnderstandingTD} and \cite{He_init_2015} paper for kernel initialization. For more detailed proof follow the \cite{He_init_2015} paper.
Let $y_l$ is the neuron output, $n_l$ is the number of input connections to the neuron, and $x_l$ is  the input to the neuron.
In a neural network (CNN/FCNN) neuron output of any $l^{th}$ layer is given as (as of now ignore the bias term): 
\begin{align}
y_l=W_lx_l
\end{align}
$x_l=y_{l-1}$, is the activation of the $(l-1)^{th}$ layer. Variance of $y_l$ is given by,
\begin{align}
Var[y_l]&=n_l Var[w_lx_l] \\
Var[y_l]&=n_lVar[w_l]E[x_l^2] 
\end{align}
Assumption is made that mean of $x_l$ and $w_l$ is zero. $E[x_l^2] = \frac{1}{2}n_lVar[y_{l-1}]$, when $f$ is ReLU.
\begin{align}
Var[y_l]&=\frac{1}{2}n_lVar[w_l]Var[y_{l-1}]\\
Var[y_L]&=Var[y_1]( \prod_{l=2}^L\frac{1}{2}n_lVar[w_l])
\end{align}
$\prod_{l=2}^L\frac{1}{2}n_lVar[w_l]$, expression will define the variance of the $y_L$, with respect to $y_1$. Make this expression as $V_A$. 
\begin{align}
\label{main_act_equ}
\prod_{l=2}^L \frac{1}{2}n_lVar[w_l]= V_A   
\end{align}
\subsection{Backward Propagation Case}
In this sub-section, we will study the backward propagation case. We will proceed with the proof given in the \cite{Glorot2010UnderstandingTD} and \cite{He_init_2015} paper for kernel initialization. For more detailed proof follow the \cite{He_init_2015} paper. The gradient of a neural network is calculated using:
\begin{align}
\Delta x_l = \hat{W_i}\Delta y_l \\
\Delta x = \frac{\partial\varepsilon}{\partial x} \\ 
\Delta y = \frac{\partial\varepsilon}{\partial y} \\
\Delta y_l = f'(y_l)\Delta x_{l+1} \\
\Delta x_i = \hat{W_i}\Delta y_i
\end{align}
Derivation of the variance of gradient reached at any layer $l$ is  as follow,
\begin{align}
Var[\Delta x_l]=\hat{n_l}Var[w_l]Var[\Delta y_{l}]
\end{align}
If activation is RElu, then expression becomes,
\begin{align}
Var[\Delta x_l] &= \frac{1}{2}\hat{n_l}Var[w_l]Var[\Delta x_{l+1}] \\
Var[\Delta x_2] &= Var[\Delta x_{L+1}](\prod_{l=2}^L\frac{1}{2}\hat{n_l}Var[w_l])
\end{align}
\begin{align}
\label{main_grad_equ}
    \prod_{l=2}^L\frac{1}{2}\hat{n_l}Var[w_l]=V_G
\end{align}
\section{Impact of Incorporating Depth of Layer on Weights Initialization}
In literature, already proposed weights initialization schemes works well, however, in practice as depth of the network increases, these initialization schemes result in training below par due to vanishing gradient problem \cite{sussillo2014random}. In this paper, we will discuss various approaches to initialize the weights by incorporating the layer's depth information and will do comprehensive study of its impact on network training. Glorot et. al. \cite{Glorot2010UnderstandingTD} and He. et. al. \cite{He_init_2015} studied the impact of initialization on the variance of activation, in forward case and variance of error gradients during back propagation. Their study suggests to initialize the weights to produce the unit variance of activation and error gradients as well. However, for deeper networks, vanishing gradient problem appears, which is addressed in this paper at some extent. Change in the variance of activations or error gradients is the function of layer's depth. In below sub-sections, we will discuss the impact of different schemes of kernel initialization, which incorporates the depth information. We carry our study giving more emphasis on backward propagation study.
\subsection{Backward/Forward Propagation Analysis}
In this sub-section, we will the study the impact of incorporating depth of each layer in weight initialization from perspective of backward propagation step. As per equation \ref{main_act_equ} and \ref{main_grad_equ}, $n_l=\hat{n_{l-1}}$. So, the analysis of backward propagation steps will also hold almost same for the forward propagation step. As per equation \ref{main_grad_equ}, assume that, $Var[w_l]=\alpha_l\beta_l$ i.e, variance of weight of each layer depends on the $\alpha_l, \beta_l$, which changes with $l$. If, as per the \cite{He_init_2015}, set $\alpha_l=\frac{2}{\hat{n_l}}$ [PUT SOME MORE DESCRIPTION ABOUT THIS ASSIGNMENT].
\begin{align}
    \prod_{l=2}^L\frac{1}{2}\hat{n_l}Var[w_l]=V_G\\
    \prod_{l=2}^L\frac{1}{2}\hat{n_l}\alpha_l\beta_l=V_G\\
    (\prod_{l=2}^L\frac{1}{2}\hat{n_l}\alpha_l)(\prod_{l=2}^L\beta_l)=V_G\\
    (\prod_{l=2}^L\frac{1}{2}\hat{n_l}(\frac{2}{\hat{n_l}}))(\prod_{l=2}^L\beta_l)=V_G\\
    (1)(\prod_{l=2}^L\beta_l)=V_G\\
    \prod_{l=2}^L\beta_l = V_G
\end{align}
In below section, we will see the impact of changing $\beta$, with respect to layer $l$. 
\subsubsection{$\beta \not\propto l$}
One way to scale the weights of each layer with any fixed constant $\beta$, then 
\begin{align}
    V_G= (\beta)^{L-1} \\
    \beta=V_G^\frac{1}{L-1}
\end{align}
We need to scale the variance of each layer weight with factor of $V_G^\frac{1}{L-1}$ to give total variance of $V_G$. $V_G > 1\implies \beta >1 $, and $V_G < 1\implies \beta <1 $. First analyse the previous approach present in literature \cite{He_init_2015}. In literature, approach is to keep the variance of each layers kernel same. As depth of the layer keeps increasing in backward propagation, intensity of vanishing gradient keeps increasing with respect to the initial value of weight $w_l$. If all weights in all layers follow the same variance $\alpha\beta$, then product of all the higher layer weight variances having variance $\alpha\beta$ will be less than the current layers variance $\alpha\beta$, given that $\alpha\beta<1$ .
\subsubsection{$\beta \propto l$ }
As depth increases, variance of activation decays gradually in forward direction and variance of error gradient decay in backward direction. This behavior suggests that, its better idea to gradually increase variance or amplify the forward signal with a factor which depends on the layer's depth. It does help in keeping the variance and total amplification of the forward signal well-tuned as depth increases. Given, $\beta = f(l)$ i.e $\beta$ scaling factor is a function of  each layers depth $l$. There may be three scenario in this setup, to reach the network variance $\leq V_G$, given that $\beta_l$ should be be in increasing order i.e, $\forall l \in L$, $\beta_l < \beta_{l+1}$. 
\begin{enumerate}
    \item \label{i1} If $\forall l \in L$, $\beta_{l-1}<\beta_{l}$, and $\beta_l < 1$, results in $V_A<1, V_G<1$.
    \item \label{i2} If $\forall l \in L$, $\beta_{l-1}<\beta_{l}$, and $\beta_l > 1$, results in $V_A>1, V_G>1$.
    \item \label{i3}  If $\forall l \in L$, $\beta_{l-1}<\beta_{l}$, and $\exists l \in L$, such that $\beta_l > 1$, and $\exists l \in L$, such that $\beta_l < 1$, results in $V_A<1, V_G<1$ or $V_A>1, V_G>1$.
\end{enumerate}
For \ref{i1}, value of $V_A$ and $V_G$ suggests that, it results in reducing the variance of subsequent layer thus de-amplifying the forward signal as well as backward signal. The lower weights also results in prone to vanishing gradient problem more.
For \ref{i2}, value of $V_A$ and $V_G$, suggests that, it results in increasing the variance of subsequent layer thus highly amplifying the forward signal. The higher weights also results in prone to exploding gradient problem more. It has been observed that very higher weights are not desirable in deep learning for better training.
For \ref{i3}, value of $V_A$, suggests that, it results in reducing or increasing the variance of subsequent layer thus de-amplifying or highly amplifying the forward signal as well as backward signal. In this case, lower layers are initialized with smaller weights, while higher layers are initialized with larger weights. We prefer to choose $f(l)$ for $\beta$, such that it results $V_A\geq 1, V_G\geq1$. There is need to obtain the gradient value after layer $l$, which is not relatively very low with the distribution of $w_l$. Distribution of all weights of layer having depth $>l$ are greater than distribution of $w_l$. The reason behind increasing weight variance proportional to depth is, top layers must have higher weights as compared to lower layers, results in eliminating the vanishing gradient problem. Higher weights of the layers on right side  make sure that it helps in mitigating the vanishing gradient phenomena, and decreasing the distribution of weights proportional to the layer depth while traversing from right to left make sure that obtained gradient at layer $l$ matches with the the weight distribution $w_l$. Therefore, it solves the problem of vanishing gradient and it increases the variance of backward signal to propagate smoothly. So increasing weight are also well in amplifying the gradients in backward direction. This approach also gives the regularization effect where concept of weight decay is implemented in the initialization scheme. In forward step, input signal is to get carried till the output layer without being weak, while in back propagation step, gradient signal is to propagate till the first layer without the loss of error gradient variance. Major significance of error-gradient is to update the weight to decrease the training loss. Gradient distribution should neither be very high nor be very low in respect with the initial distribution of the weight to be updated. Very low gradient makes training very slow and stops the training, while very high gradient results in larger weight update, which causes unstable training. Let understand the back propagation step in deep learning. This can also be understood, as higher variance helps in preventing the signal from decaying. One main thing to notice in this approach is, while back propagation step, gradients amplify gradually. Let us understand it mathematically; at layer $m$, weights are $w$. Right hand product of weights are higher than, weights without scaled. However, this product value does gradually down during back propagation. In addition, at layer $m$, it matches with the distribution of weights at $m^{th}$ layer. This property is favorable in training, because, if product does not match with the distribution of the weights, then comparatively higher weight updates can lead the training to overfit or underfit regime.
\subsubsection{$\beta \propto \frac{1}{l}$}
There may be three scenario in this setup, to reach the network variance $\leq V_A$ and value of $\beta_l$ should be be in decreasing order. 
\begin{enumerate}
    \item \label{i1} If $\forall l \in L$, $\beta_{l}<\beta_{l-1}$, and $\beta_l < 1$, results in $V_A<1$.
    \item \label{i2} If $\forall l \in L$, $\beta_{l}<\beta_{l-1}$, and $\beta_l > 1$, results in $V_A>1$.
    \item \label{i3}  If $\forall l \in L$, $\beta_{l}<\beta_{l-1}$, and $\exists l \in L$, such that $\beta_l > 1$, and $\exists l \in L$, such that $\beta_l < 1$, results in $V_A>1$ or $V_A<1$.
\end{enumerate}
For \ref{i1}, value of $V_A$, suggests that, it results in reducing the variance of subsequent layer thus de-amplifying the forward signal. The lower weights also results in prone to vanishing gradient problem more.
For \ref{i2}, value of $V_A$, suggests that, it results in increasing the variance of subsequent layer thus highly amplifying the forward signal. The higher weights also results in prone to exploding gradient problem more. It has been observed that very higher weights are not desirable in deep learning.
For \ref{i3}, value of $V_A$, suggests that, it results in reducing of increasing the variance of subsequent layer thus de-amplifying or highly amplifying the forward signal. In this case, initial layers are initialized with lower weights while higher layers are initialized with higher weights.
From the variance perspective, this approach must also works well as with the previous approach, however this is not true. This approach is not performing well because of lower weights in top layers and comparatively higher weights in lower layers. Decaying weights in top layers cause vanishing gradient problem. This setup of weight cause the vanishing gradient problem at higher extent. This is shown empirically in the figure abc as well.
\section{Depth-wise Increased Scale Initialization}
While making decision regarding which initialization scheme to pick, there is also need for detailed discussion on which activation function is applied on that layer. Activation function plays significant role in initialization scheme. In \cite{Glorot2010UnderstandingTD}, study is made on the sigmoidal activation in general, while on specifically on Tanh activation. In \cite{He_init_2015}, comprehensive study on the ReLU activation. In this paper, we will do detailed study initialization scheme from the perspective of rectified linear units. Proposed weight initialization scheme can be understand by analysing how weights at specific depth is affecting the network training. One notable observation related to weights is, there exist the vanishing gradient problem in back-propagation step, same as vanishing activation problem also exist in forward pass step. Common practice in weight initialization scheme is, weights are initialized with small values, less than one. Therefore, when activation value propagates further in the layer then its value start vanishing. That is why, it is better approach to initialize earlier layer with lower weights and keep increasing the weights in subsequent layers. One question arises, if high weights are assigned to top layers, then what will be its effect. There is much deeper explanation is needed on this. If layer number is passing in increasing order then it is giving very great result while if layer number is passing in decreasing order then it gives result  below par.
However, when $n$ hidden layers use an activation like the sigmoid function, $n$ small derivatives are multiplied together. Thus, the gradient decreases exponentially as we propagate down to the initial layers. A small gradient means that the weights and biases of the initial layers will not be updated effectively with each training session. Since these initial layers are often crucial to recognizing the core elements of the input data, it can lead to overall inaccuracy of the whole network.
\section{Introducing Depth-wise Initialization Function}
$\beta_l$ is the scaling factor at $l_{th}$ layer. Let introduce, a function $\alpha^{\left( \frac{1}{log_K(l)}-1\right)}$ which is the function of layers depth $l$, which will be used to scale the weights depth-wise and will be work as $\beta_l$.
\begin{align}
\beta_l &= \alpha^{\left( \frac{1}{log_K(l)}-1\right)}
\end{align}
By putting value of $\beta_l$ in equation \ref{main_grad_equ}.
\begin{align}
\prod_{l=2}^L\frac{1}{2}n_l\alpha*\alpha^{\left( \frac{1}{log_K(l)}-1\right)}&=V \\
\prod_{l=2}^L\frac{1}{2}n_l\alpha^\frac{1}{log_K(l)}&=V \\
(\frac{1}{2})^{L-1}(\prod_{l=2}^L n_l)\alpha^{\sum_{l=2}^L\frac{1}{log_K(l)}}&=V \\
\alpha^{\sum_{l=2}^L\frac{1}{log_K (l)}}&=V\frac{2^{L-1}}{(\prod_{l=2}^L n_l)}\\
\sum_{l=2}^L\frac{1}{log_k(l)}&=log_\alpha \left(V\frac{2^{L-1}}{(\prod_{l=2}^L n_l)}\right)
\end{align}
Let assume that, $n_l=n$, is same for all the hidden layers.\\
\begin{align}
\sum_{l=2}^L\frac{1}{log_K(l)}&=log_\alpha V + log_\alpha(\frac{2}{n})^{L-1} \\
\sum_{l=2}^L\frac{1}{log_K(l)}&=log_\alpha V + (L-1)log_\alpha\frac{2}{n} \\
\sum_{l=2}^Llog_l K&=log_\alpha V + (L-1)log_\alpha\frac{2}{n}
\end{align}
In above expression, two values $\alpha$ and $K$ need to be determined. For sake of easy calculation, take $\alpha=\frac{2}{n}$.\\
\begin{align}
\sum_{l=2}^Llog_l K&=log_\frac{2}{n} V + (L-1)log_\frac{2}{n}\frac{2}{n}\\
\label{derived_equation}\sum_{l=2}^Llog_l K&=log_\frac{2}{n} V + (L-1)   
\end{align}
\section{Initialization function analysis}
$log$ is used just to make sure that higher level weights does not increase in greater amount, however lowering the value of lower layer weights does not hurt. If $V>1$, then $log_\frac{2}{n}V<0$, and for $V<1$, then $log_\frac{2}{n}>0$, and $V\leq 0$ is not valid. One notable thing about $\alpha < 0$, keeping value of $V$ constant is, higher value of $\alpha$  cause higher value of $K$, on contrary, lower value of $\alpha$ causes lower value of $K$. How to decide what variance should be of the given network. This depends on network depth and data in hand. Variance of the network can be managed by adjusting value of $K$. It works as a hyper parameter for the training of the network. Major problem in deep learning is to amplifying the activation outputs from initial state to last state and carrying of gradient information from last state to initial state. Proposed approach solves the above problem in proper way. What variance is appropriate for your training depends on several factors. However, this approach provides a flexibility to increase and decrease the variance according to the training requirements. For deeper layer high variance is required while for shallow layers lesser variance is required. It depends on practitioner what layer depth he picks.
\section{Hyper-parameter analysis}
Using equation in \ref{derived_equation}, for $V=1$,
\begin{align}
\sum_{l=2}^Llog_l K &=log_\frac{2}{n}(1) + (L-1)\\
\sum_{l=2}^Llog_l K &= (L-1) 
\end{align}
For $V{\not=}1$,\\
\begin{align}
\sum_{l=2}^Llog_l K=log_\frac{2}{n} V + (L-1)\\
\sum_{l=2}^L\frac{log_e K}{log_e l}=\frac{log_e V}{log_e\frac{2}{n}} + (L-1)\\
log_eK\sum_{l=2}^L\frac{1}{log_e l}=\frac{log_e V}{log_e\frac{2}{n}} + (L-1)\\
\frac{(L-1)}{log_eL} <\sum_{l=2}^L\frac{1}{log_el} < \frac{(L-1)}{log_e2}\\
\sum_{l=2}^L\frac{1}{log_el} 	\approx \frac{L-1}{log_ef(l)}\\
(L-1)*log_eK\frac{1}{log_ef(l)}=\frac{log_e V}{log_e\frac{2}{n}} + (L-1)\\
log_eK = \frac{log_e V}{log_e\frac{2}{n}}\frac{log_ef(l)}{(L-1)} + log_ef(l)\\
log_eK = log_ef(l)(\frac{log_e V}{log_e\frac{2}{n}}\frac{1}{(L-1)} + 1)\\
K=f(l)^{(\frac{log_e V}{log_e\frac{2}{n}}\frac{1}{(L-1)} + 1)}\\
f(l)<1\\
K\propto \frac{L}{V*n}
\end{align}
It implies that increasing the value of $V$ and $n$ will decrease the value of $K$, while decreasing the value of $L$ will also decrease the value of $K$.
if $V>1$, then $log_\frac{2}{n}V<0$, and for $V<1$, then $log_\frac{2}{n}>0$, and $V\leq 0$ is not valid. In above expression, it can seen that $K \propto \frac{1}{V}$. Low value of $K$ leads to higher weights. One notable thing about $\alpha < 0$, keeping value of $V$ constant is, higher value of $\alpha$  cause higher value of $K$, on contrary, lower value of $\alpha$ causes lower value of $K$. Variance of the network can be managed by adjusting value of $K$. It works as a hyper parameter for the training of the network.
\paragraph{}
So, one observation is, initial weight distribution of increasing higher variance, or distributed mean does not affect the performance of the learning, however once model reaches in higher iterations, problem starts of converging due to the vanishing gradient problem and higher variance problem. To mitigate the problem of higher variance, apply the batch normalization. One significant observation is, not very high weights are desirable. there must be an analysis over how many layers should be inside the one scale and how many out of the one scale. it has been observed that, 33 percent weights should be below one scale and rest are at above one. To make the layers below one scale, $K$ should be higher than all those layer depths, and vice-versa is also true. To make the adjustment between $K$ and layer add some constant in the base i.e $l$ in $log_lK$. Some constant needs to be added in base just to make sure that scale should not be very low, because for initial layers i.e 2 to 7 value of $log_lK$ will became very low. To avoid this we can do $log_{(l+constant)}K$. It is observed that lower weights are favourable to quickly converge on training and testing(not always), however it then cause for vanishing gradient, there is need to find the optimum, where optimum is achieved to make a combination between lower and higher weights.
\section{Experiment}
The experiment is conducted on the relu activation. More experiment need to carry on tanh activation and pexpo activation. Carry experiment on number of layers, Number of hidden units per layer. Normal, uniform. Convolution network cifar10. Fcnn mnist. Increasing, decreasing, same, unit variance. Increasing, decreasing, same under the regime of keeping unit variance.
Show accuracy curve, loss curve, Weight distribution, gradient distribution, activation distribution
Use history to draw the learning curve. Use tensorboard for gradient weight and activation distribution. Things to keep in mind f
\begin{figure}[t]
    \centering
    \includegraphics[scale=0.6]{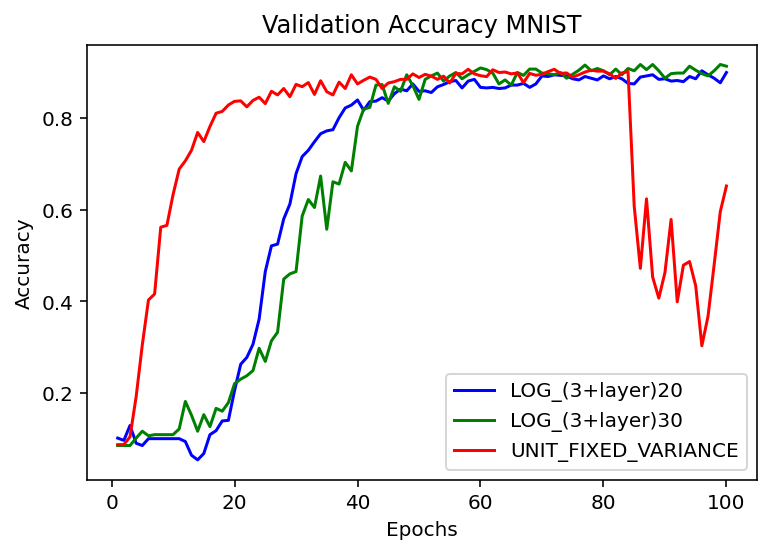}
    \caption{Caption}
    \label{fig:my_label}
        \centering
    \includegraphics[scale=0.6]{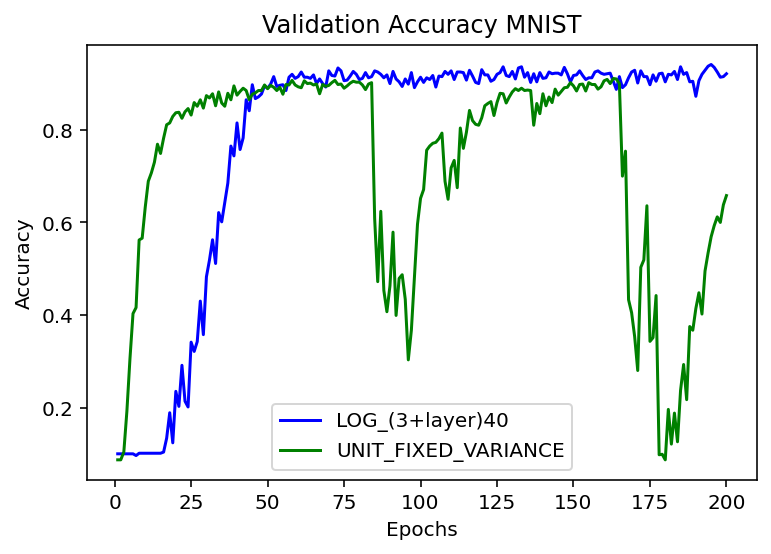}
    \caption{Caption}
    \label{fig:my_label2}
\end{figure}
Experiments are carried on cifar10 dataset with very deep fcnn of 54 layers and 64 neurons each. In terms of initializer, both normal and uniform distribution is used. Four cases are considered:He and Xavier initialization scheme is used: With unit variance, With 22 variance and increasing layer wise weight variance, With 22 variance and decreasing layer wise weight variance, With 22 variance and constant various across all layers.
\section{Discussion}
Main point came out of above discussion is, applying knowledge of depth in initialization has significant role in the training. Top layer weights are directly exposed to loss function. Any change in weight directly affects the loss function. On contrary, change in weights at the first layer have indirect small impact on loss function.
One question arises is, whether it is better to give higher feature importance to lower layer activations, or activations at top layer. Our experiment suggests that, it is better to provide smaller weights to lower layers as compared to top layers, reason behind this can be understood in this way; suppose, if less importance is given to each feature because there is no understanding of which is important or less important. During training in forward propagation, higher layer weights will be trained in such a manner that important features will get higher weights in top layer features while less important features will be ruled out while higher importance features will attain higher weights. One of the critical finding in the carried experiment is; if distribution is uniform in the he initialization scheme, the training stalls and does not progress at all, while if distribution is normal then training progresses very well. Meaningful observation is, proposed method works well for uniform distribution as well as for normal and does not stall with any distribution. One key thing to mention here is normal distribution is superior over uniform distribution in this initialization setup.
Why he initialization scheme does stall for uniform distribution while works for normal distribution.
All of the above variance based experiment works on the assumption that input data is normalized to zero. However, in our experiments we scaled the data between 0 and 1 and not normalized it. One intuitive way to understand this is from weight distribution perspective.
\section{Conclusion}
\bibliography{ms.bib}
\bibliographystyle{plain}
\end{document}